\documentclass[pdflatex,sn-mathphys-num]{sn-jnl}

\usepackage{graphicx}
\usepackage{multirow}
\usepackage{amsmath,amssymb,amsfonts}
\usepackage{amsthm}
\usepackage{mathrsfs}
\usepackage[title]{appendix}
\usepackage{xcolor}
\usepackage{textcomp}
\usepackage{tikz}
\usetikzlibrary{arrows.meta, positioning, calc, shapes.geometric, arrows}
\usepackage{booktabs}
\usepackage{algorithm}
\usepackage{algorithmicx}
\usepackage{algpseudocode}
\usepackage{listings}
\usepackage{multirow}
\usepackage{pifont}
\usepackage{url}
\usepackage{soul}

\DeclareMathOperator*{\argmin}{arg\,min}

\newcommand{\cmark}{\ding{51}}
\newcommand{\xmark}{\ding{55}}

\theoremstyle{thmstyleone}

\theoremstyle{thmstyletwo}

\theoremstyle{thmstylethree}

\begin{document}

\title[DiaSeg]{DiaSeg: Diagonal Segment Extraction from DTW Paths
for Interpretable Gait Analysis}

\author*[1]{\fnm{Tresor Y.} \sur{Koffi}}
\email{tresor.koffi@u-bourgogne.fr}

\author[1]{\fnm{Amel} \sur{Hidouri}}
\email{amel.hidouri@u-bourgogne.fr}

\author[1]{\fnm{Corentin} \sur{Legrand}}
\email{corentin\_legrand@etu.ube.fr}

\author[1]{\fnm{Aurélie} \sur{Bertaux}}
\email{aurelie.bertaux@u-bourgogne.fr}

\affil*[1]{\orgname{Université Bourgogne Europe},
\orgdiv{CIAD UR 7533},
\orgaddress{\city{Dijon}, \postcode{21000}, \country{France}}}

\abstract{Dynamic Time Warping (DTW) is the dominant approach
for measuring similarity between time series, yet standard
practice discards the optimal warping path after computing a
single distance value, losing precisely the local alignment
information most relevant to clinical diagnosis.
We introduce DiaSeg, a framework that recovers this
discarded information by extracting diagonal segments from DTW
paths with controlled breaks, characterizing each segment by
five geometric features, and
enabling unsupervised pattern discovery without domain-specific
feature engineering. Validated on 91 subjects across six
clinical conditions (healthy aging, Parkinson's, Huntington's,
ALS, brain tumor, and stroke), three findings emerge. First,
diagonal segments form consistent unsupervised patterns
(silhouette 0.33) aligned with biomechanical phase annotations,
with external validation confirming near-perfect separation of
healthy and pathological gait (ARI up to 0.986). Second,
segments discriminate pathology at 69\% (supervised) and 75\%
(patient-level clustering), with pathology manifesting through
distributional shifts in segment length rather than individual
segment properties; combining segment and cycle-level features
further improves classification to 91.7\%. Third, while
cycle-based methods achieve higher accuracy (91\%), diagonal
segments provide phase-specific interpretability unavailable
in global representations, localizing where coordination
breaks down within the gait cycle. DiaSeg thus transforms DTW
from a black-box distance into a source of interpretable
temporal features for neurodegenerative disease assessment.}
\maketitle

% ====================================================================
\section{Introduction}
\label{sec:introduction}
% ====================================================================

Clinical movement analysis increasingly relies on time series
comparison to detect and monitor neurodegenerative disorders,
where subtle deviations in gait rhythm, coordination, and temporal
structure can signal disease onset or progression long before they
become clinically apparent~\cite{aghabozorgi2015time}. Beyond
healthcare, time series analysis has become essential across
scientific and engineering domains, from industrial process control
to financial forecasting and environmental sensing. A fundamental
challenge across all these domains is measuring similarity between
temporal sequences that may exhibit different speeds, phases, or
temporal distortions while sharing underlying patterns. Dynamic
Time Warping (DTW)~\cite{sakoe1978dynamic} addresses this challenge
by computing an optimal non-linear alignment between sequences,
making it one of the most widely adopted techniques in time series
classification~\cite{keogh2001derivative},
clustering~\cite{petitjean2011global}, and
retrieval~\cite{keogh2005exact}.
Traditional DTW applications treat the algorithm as a black box that
consumes two sequences and produces a scalar distance
measure~\cite{keogh2005exact,salvador2007toward}. Research has
primarily focused on two directions, namely improving computational
efficiency through lower bounds~\cite{keogh2005exact} and approximate
algorithms~\cite{salvador2007toward}, and refining distance computation
through constrained search strategies~\cite{ratanamahatana2004making}.
While effective for similarity measurement, both directions share a
fundamental limitation in that the optimal warping path, which
encodes the sequence of alignment decisions defining how temporal
features correspond between series, is computed and then discarded.
In clinical contexts, this represents not merely a technical
inefficiency but a meaningful loss of diagnostic information,
specifically which phases of movement align well with a healthy
reference, where coordination breaks down, and how sustained
synchrony degrades with disease progression.
Recent advances in time series analysis have explored alternative
approaches to pattern discovery. Shapelet-based
methods~\cite{ye2009time} identify discriminative subsequences for
classification but require labeled data and focus on discriminative
rather than alignment patterns. Recurrence Quantification Analysis
(RQA)~\cite{marwan2007recurrence,webber2015recurrence} analyzes
diagonal structures in recurrence plots by comparing all time points
against fixed thresholds, successfully characterizing dynamical
systems in physiology~\cite{marwan2002recurrence} and
geosciences~\cite{marwan2002cross}. However, RQA operates on
recurrence matrices fundamentally different from DTW's optimal
alignment paths and addresses state-space recurrence rather than
optimal temporal correspondence. Deep learning
approaches~\cite{wang2017time} automatically learn representations
but lack interpretability and require large labeled datasets,
limiting their applicability to unsupervised pattern discovery in
domains with limited clinical data. Thus, despite these advances,
systematically extracting and quantifying local alignment patterns
from DTW optimal paths for unsupervised clustering remains unexplored.
When two time series share similar temporal patterns, the DTW
path exhibits characteristic geometric structures, particularly
diagonal segments, which are contiguous portions progressing
diagonally through the cost matrix, indicating sustained temporal
synchrony where both sequences advance together. However,
real-world time series, especially from biological or physical
processes, rarely exhibit perfect synchrony. Minor temporal
irregularities, measurement noise, and natural variability introduce
occasional non-diagonal steps within otherwise well-aligned regions.
A rigid extraction of only strict diagonal segments would fragment
these alignment regions and fail to capture sustained correspondence,
while existing methods provide no principled mechanism to balance
strictness with robustness in identifying temporal patterns.
To address these challenges, we propose DiaSeg, a framework that
systematically extracts and quantifies diagonal segments from DTW
optimal paths, tolerating a controlled number of non-diagonal steps
to capture sustained yet flexible temporal correspondences. By
allowing breaks within otherwise diagonal segments, DiaSeg extracts
meaningful alignment patterns that balance strictness with robustness,
identifying sustained temporal correspondences while accommodating
minor desynchronizations inherent in noisy data. We characterize each
segment through five features namely effective length $L$, break count
$\ell$, cost variation $d$, temporal position $t_0$, and path context
$p_l$, enabling unsupervised pattern discovery through clustering
without requiring predefined features or domain expertise. This
approach transforms DTW from a scalar distance measure into a rich
source of structural information for data-driven temporal pattern
mining in clinical movement analysis.
The main contributions of this paper are as follows:
\begin{itemize}
    \item We propose an efficient algorithm to extract diagonal
segments from DTW paths with controlled break tolerance,
enumerating maximal segments for each admissible break count
up to a user-defined maximum $\ell_{\max}$, with sensitivity
analysis confirming stable clustering quality across a range
of break tolerance values.

    \item We define five complementary features characterizing each
segment, namely effective length, break count, cost variation,
temporal position, and path context, encoding geometric, temporal,
and cost-based alignment properties in a unified representation.

   \item We introduce a clustering-based framework that aggregates
segment features across pairwise DTW comparisons, enabling
data-driven temporal pattern discovery without requiring
domain-specific feature engineering or labeled data,
with label-based validation confirming near-perfect
healthy--pathological separation (ARI up to 0.986).

    \item We provide comprehensive validation on three independent
datasets, including ground truth biomechanical phase labels (BLISS),
demonstrating disease-specific temporal fragmentation signatures
and pathology discrimination at 69\% (supervised) and 75\%
(patient-level clustering); we further show that combining
segment and cycle-level features improves classification to 91.7\%,
establishing empirical complementarity between local and global
gait representations.
\end{itemize}

The remainder of this paper is organized as follows.
Section~\ref{sec:related_work} reviews related work in DTW-based
methods, time series clustering, recurrence quantification
analysis, and warping path analysis, positioning our
contributions within the broader literature.
Section~\ref{sec:methodology} presents the core methodology,
including problem formulation, DTW computation, the diagonal
segment extraction algorithm with formal definitions and
pseudocode, feature quantification, and the unsupervised
clustering framework. Section~\ref{sec:experiments} reports experimental validation, including unsupervised
clustering on gait patterns, supervised discrimination analysis
across multiple pathologies, and ground truth validation using
phase-labeled data. Section~\ref{sec:conclusion} concludes the
paper and outlines future research directions.

% ====================================================================
\section{Related Work}
\label{sec:related_work}
% ====================================================================

Dynamic Time Warping (DTW) has established itself as a
technique for temporal sequence comparison since its introduction
by Sakoe and Chiba~\cite{sakoe1978dynamic}. While DTW-based
similarity measures and their variants are widely utilized across
diverse applications, the analytical exploitation of the optimal
warping path's structural properties remains underexplored. This
section reviews existing approaches to time series analysis,
clustering, and pattern discovery, highlighting the limitations
that motivate our work.

\subsection{DTW and Its Computational Variants}

DTW computes an optimal alignment between two temporal sequences by
finding a warping path that minimizes cumulative distance under
monotonicity and boundary constraints~\cite{sakoe1978dynamic}. Its
ability to handle non-linear temporal distortions has made it
invaluable for time series classification~\cite{keogh2001derivative}
and clustering~\cite{petitjean2011global}. Research has primarily
evolved along two directions, namely improving computational
efficiency and refining distance computation. On the efficiency
side, lower bounding techniques~\cite{keogh2005exact}, approximate
methods~\cite{salvador2007toward}, and constrained search
strategies~\cite{ratanamahatana2004making} have been proposed to
address DTW's quadratic time complexity. Among these, the
Sakoe-Chiba band~\cite{sakoe1978dynamic} limits the warping path
to a diagonal corridor, reducing computation and preventing
pathological alignments where single points map to large
subsequences.
On the distance refinement side, numerous variants have emerged
over the past two decades. Keogh and Pazzani~\cite{keogh2001derivative}
introduced Derivative DTW (DDTW), aligning first-order derivative
features rather than raw values to capture shape similarity over
magnitude matching. Jeong et al.~\cite{jeong2011weighted} proposed
Weighted DTW (WDTW), applying phase-difference-based penalties to
reduce pathological alignments. Batista et al.~\cite{batista2014cid}
developed the complexity-invariant distance (CID) to handle
differences in series complexity through a correction factor.
Zhao and Itti~\cite{zhao2018shapedtw} introduced Shape DTW using
local shape descriptors including discrete wavelet transforms and
histograms of gradients, while Yuan et al.~\cite{yuan2019locally}
proposed Locally Slope DTW (LSDTW) incorporating local slope
features with filtering techniques.
Despite this diversity, a comprehensive comparative study by
Lahreche and Boucheham~\cite{lahreche2021comparison} evaluating
these variants on 85 datasets from the UCR archive found no
statistically significant performance differences for classification
tasks, suggesting that incremental modifications to distance
computation have reached diminishing returns. More fundamentally,
all these variants share the same critical limitation as classical
DTW, in that regardless of whether they modify local distance
functions, apply constraints, use derivative features, or
incorporate complexity measures, they all produce a single scalar
distance value and universally discard the optimal warping path
after extracting that value. This systematic neglect of path
structure represents a fundamental
gap, leaving the rich local alignment information encoded in the
warping path almost entirely unexploited.

\subsection{Time Series Clustering Approaches}

Unsupervised clustering of time series typically relies on
distance-based methods such as k-means with DTW
distance~\cite{petitjean2011global} or hierarchical
clustering~\cite{aghabozorgi2015time}. These approaches use global
similarity measures, whether scalar DTW distances or Euclidean
distances, to partition sequences into groups sharing similar
overall patterns. Petitjean et al.~\cite{petitjean2011global}
proposed DTW Barycenter Averaging (DBA) for computing centroids in
DTW space, enabling more effective k-means clustering with DTW.
Model-based techniques, including Gaussian Mixture Models
(GMM)~\cite{reynolds2009gaussian}, have been successfully applied
to various domains, enabling probabilistic assignment of sequences
to clusters with uncertainty quantification. While effective for
many applications, these methods operate on scalar distance measures
or hand-crafted features, failing to exploit the local alignment
structure inherent in DTW comparisons.
In gait analysis specifically, Kim et al.~\cite{kim2023one} used
global DTW distances to classify hip osteoarthritis patients from
healthy controls using ground reaction forces. Park
et al.~\cite{park2025instant} employed GMM on manually engineered
correlation and error metrics derived from force plate data. Ewen
et al.~\cite{ewen2021identification} performed k-means clustering
on principal components derived from gait kinematics, identifying
patient subgroups based on peak hip extension and other kinematic
features. Abdul Halim et al.~\cite{abdul2022cluster} combined PCA
with k-means on biomechanical features and pain scores. Despite
their success in identifying patient subgroups, these approaches
share two critical limitations.
First, they require domain-specific feature engineering and manual
selection of relevant features, limiting generalizability across
application domains and demanding expert knowledge for each new
problem. Second, when DTW is used, only the final scalar distance
enters the clustering algorithm, while the path revealing where and
how sequences align or misalign is discarded. This represents a
fundamental information loss, as the path encodes local alignment
quality, phase-specific synchrony, and temporal correspondence
patterns that could distinguish subtypes invisible to global
distance measures alone. DiaSeg addresses both limitations by
operating K-means not on scalar distances or hand-crafted features
but on structural features derived directly from the warping path
geometry, so that the clustering algorithm organizes alignment
structure rather than bypassing it.

\subsection{Shapelet-Based Pattern Discovery}

Shapelet-based methods represent an alternative approach to time
series pattern discovery. Ye and Keogh~\cite{ye2009time} introduced
shapelets as discriminative subsequences learned from labeled data
for classification tasks. Shapelets identify local patterns that
distinguish between classes, making them valuable for supervised
learning scenarios where the goal is to maximize class separation \cite{rakthanmanon2013fast}.
The approach has been extended with various learning algorithms and
has achieved competitive classification performance on benchmark
datasets \cite{zhang2021elis++} \cite{jing2024method}.
However, shapelet discovery is fundamentally designed for supervised
classification rather than unsupervised clustering \cite{cai2024se}. Shapelets require
labeled training data and seek discriminative patterns that maximize
inter-class differences, whereas our objective is to discover shared
alignment patterns across unlabeled sequences in an unsupervised
manner. Furthermore, shapelets identify subsequences within
individual time series, while our approach analyzes the pairwise
alignment structure between sequences, a fundamentally different
problem requiring different methodological frameworks. In essence,
shapelets address the question of which local patterns distinguish
one class from another, whereas DiaSeg addresses how two sequences
progress through time together, capturing the phases during which
they advance in synchrony and the points at which that synchrony
breaks down. These distinct objectives make shapelet-based approaches
unsuitable for discovering temporal correspondence patterns from
pairwise DTW alignments.

\subsection{Recurrence Quantification Analysis}

A conceptually related approach to analyzing structural patterns in
temporal data is Recurrence Quantification Analysis (RQA),
introduced by Eckmann et al.~\cite{eckmann1987recurrence} and
systematically developed by Marwan and
colleagues~\cite{marwan2007recurrence,webber2015recurrence}. RQA
analyzes diagonal line structures in recurrence plots where a point
$(i,j)$ is marked if states $x_i$ and $x_j$ are sufficiently
similar according to a fixed threshold. The length and distribution
of diagonal lines quantify deterministic behavior and predictability
through measures such as determinism (DET), average diagonal line
length, and laminarity (LAM)~\cite{zbilut1992embeddings,webber1994dynamical}.
These measures have been successfully applied to characterize
dynamical systems in physiology~\cite{marwan2002recurrence},
geosciences~\cite{marwan2002cross}, and other domains. While RQA and our approach both analyze diagonal structures, they
differ fundamentally in their mathematical foundations and
objectives. RQA constructs a recurrence matrix by comparing all
state pairs $(x_i, x_j)$ against a fixed threshold $\epsilon$,
marking $R(i,j) = 1$ if $\|x_i - x_j\| < \epsilon$, thereby
identifying regions where a single time series revisits similar
states in its phase space. In contrast, our method analyzes the
single optimal path returned by DTW, a continuous valued, optimized
alignment path that represents the best temporal correspondence
between two distinct sequences under dynamic programming constraints.
The diagonal segments we extract from DTW paths capture phases of
sustained temporal correspondence between two sequences during
optimal alignment, which is conceptually and mathematically distinct
from the state-space recurrence patterns analyzed by RQA.
Furthermore, RQA's threshold parameter $\epsilon$ requires manual
specification and significantly affects results, while DTW's optimal
path is determined algorithmically without free parameters beyond
the choice of local distance function.

\subsection{Warping Path Analysis and Feature Extraction}

Despite DTW's ubiquity and the extensive research on distance
computation variants, explicit analysis of the warping path's
geometric structure for feature extraction remains remarkably rare.
Park et al.~\cite{park2006novel} introduced segmental DTW for
speaker segmentation, partitioning constrained diagonal paths into
fragments colored by average distortion to discriminate between
speakers. Their focus was on supervised speaker discrimination in a
specific domain rather than unsupervised pattern clustering, and
they did not formalize diagonal segment extraction with break
tolerance or apply it to general pattern discovery problems. Silva
et al.~\cite{silva2016speeding} studied efficient computation of
the full DTW distance matrix for all-pairwise comparisons but did
not explore path-based feature extraction for pattern discovery,
focusing purely on computational optimization.
More broadly, the machine learning community has emphasized
learning-based approaches such as deep neural
networks~\cite{wang2017time} over hand-crafted path-derived
features. While deep learning methods automatically learn
representations and have shown good results on benchmark datasets,
they provide no explicit alignment information or interpretability
regarding temporal correspondences.
Moreover, Shahcheraghi et al.~\cite{shahcheraghi2022mplots}
introduced Mplots, a scalable approach to computing and
visualising time series self-similarity matrices that reveals
recurrent patterns across massive datasets. While conceptually
related in exploiting the structure of pairwise comparison
matrices, Mplots operate on self-similarity matrices to discover
motifs within a single long time series, whereas DiaSeg analyses
the optimal alignment path between two distinct sequences to
extract geometric features of their temporal correspondence.
Hence, the comparative study by Lahreche and
Boucheham~\cite{lahreche2021comparison} provides compelling
evidence that path analysis represents an unexplored research
direction. After evaluating classical DTW and its most popular
variants on 85 datasets, they found no statistically significant
differences between virtually all variants for classification,
suggesting that refinements to distance computation have reached
saturation. To our knowledge, no prior work has systematically
extracted diagonal segments with controlled breaks from DTW optimal
paths as features for unsupervised clustering, leaving the rich
structural information in warping paths almost unexploited for
pattern discovery. Our work addresses this gap by introducing the
first systematic framework for extracting and quantifying structural
features from DTW optimal paths, transforming DTW from a black-box
distance measure into a rich source of interpretable structural
information for data-driven temporal pattern mining.

% ====================================================================
\section{Methodology}
\label{sec:methodology}
% ====================================================================

This section presents our framework for extracting and quantifying
local alignment patterns from DTW optimal paths. We first formulate
the problem, then detail the core algorithmic contributions, namely
DTW computation, diagonal segment extraction with controlled breaks,
feature quantification, and the unsupervised clustering framework.

\subsection{Problem Formulation}

\begin{figure}[H]
\centering
\begin{tikzpicture}[
    node distance=1cm,
    stage/.style={rectangle, draw, rounded corners,
                  minimum width=5.5cm, minimum height=0.9cm,
                  align=center, font=\small, fill=blue!5},
    param/.style={ellipse, draw, fill=yellow!10,
                  minimum width=1.6cm, minimum height=0.7cm,
                  align=center, font=\small},
    arrow/.style={-{Stealth[scale=1]}, thick},
    every node/.style={align=center}
]
\node[stage] (s1) {Time Series $\mathcal{X} = \{X_1,\ldots,X_N\}$};
\node[stage, below=0.7cm of s1] (s2)
    {Pairwise DTW $\mathcal{P}_{ij} = \text{DTW}(X_i,X_j)$};
\node[stage, below=0.7cm of s2, fill=orange!20] (s3)
    {\textbf{Diagonal Extraction} $\mathcal{S}$ with $\ell \leq \ell_{\max}$};
\node[stage, below=0.7cm of s3] (s4)
    {Feature Quantification $\mathbf{f} = (L,\ell,d,t_0,p_l)$};
\node[stage, below=0.7cm of s4] (s5)
    {Unsupervised Clustering $\mathcal{C}_1,\ldots,\mathcal{C}_k$};
\node[stage, below=0.7cm of s5] (s6)
    {Pattern Discovery / Clinical Interpretation};
\draw[arrow] (s1) -- (s2);
\draw[arrow] (s2) -- (s3);
\draw[arrow] (s3) -- (s4);
\draw[arrow] (s4) -- (s5);
\draw[arrow] (s5) -- (s6);
\node[param, right=0.7cm of s3] (p1) {$\ell_{\max}$};
\node[param, right=0.7cm of s5] (p2) {$k$, $\epsilon$};
\draw[arrow, dashed] (p1) -- (s3);
\draw[arrow, dashed] (p2) -- (s5);
\end{tikzpicture}
\caption{DiaSeg framework workflow. The orange-highlighted stage
represents the core contribution of extracting diagonal segments
with controlled breaks ($\ell \leq \ell_{\max}$).}
\label{fig:architecture_compact}
\end{figure}
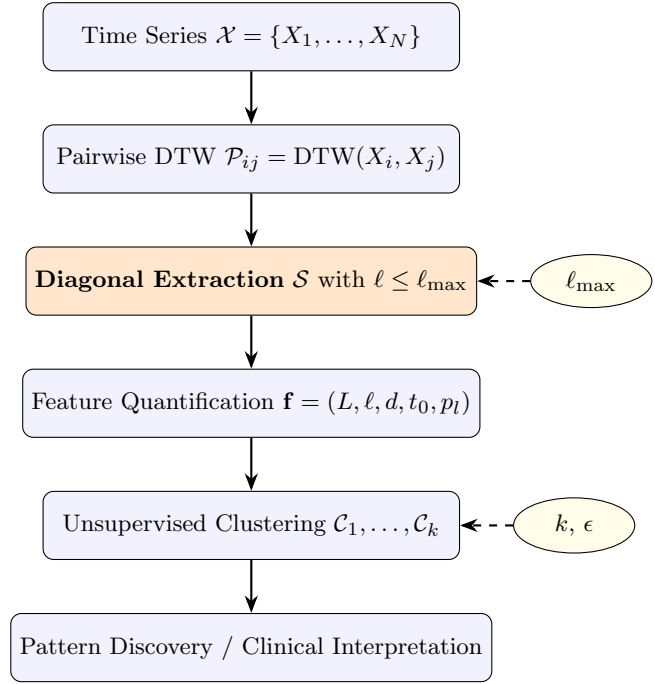

Given a collection of $N$ temporal sequences
$\mathcal{X} = \{X_1, X_2, \ldots, X_N\}$ where each sequence
$X_i = (x_1^i, x_2^i, \ldots, x_{n_i}^i)$ represents a
time-dependent observation such as gait cycles, physiological
signals, or sensor measurements, our objective is to discover
groups of sequences sharing similar temporal patterns without
requiring predefined features or labels. Traditional approaches
apply DTW to compute pairwise distances $\text{DTW}(X_i, X_j)$
and perform clustering based on these scalar similarity measures.
However, this discards the rich structural information encoded in
the optimal warping path, specifically which portions of the
sequences exhibit sustained temporal correspondence and which
require temporal adjustments.
Our key insight is that the DTW optimal warping path
$\mathcal{P}_{ij}$ between sequences $X_i$ and $X_j$ contains
geometric structures that encode local alignment quality. In
particular, diagonal segments are contiguous portions where the
path progresses diagonally through the cost matrix, indicating
regions of sustained temporal synchrony where both sequences
advance together. The challenge lies in defining a flexible yet
principled extraction mechanism that captures sustained alignment
regions while accommodating minor temporal irregularities
inherent in real-world data. We address this through
diagonal segments with controlled breaks, detailed in
Section~\ref{sec:extraction}.
Figure~\ref{fig:architecture_compact} illustrates the
complete pipeline from raw time series input through
pairwise DTW computation to clinical pattern discovery.

\subsection{Dynamic Time Warping and Optimal Path Extraction}

This section presents the mathematical foundation of DTW,
covering the cumulative cost matrix computation
(Section~\ref{sec:dtw_matrix}) and the backtracking procedure
used to recover the optimal warping path
(Section~\ref{sec:dtw_path}), which serves as the direct
input to the diagonal segment extraction framework described
in Section~\ref{sec:extraction}.

\subsubsection{DTW Cumulative Cost Matrix}
\label{sec:dtw_matrix}

Let $X = (x_1, x_2, \ldots, x_n)$ and $Y = (y_1, y_2, \ldots, y_m)$
be two temporal sequences where each $x_i, y_j \in \mathbb{R}^d$
supports multivariate signals. We define the local distance function
as the Euclidean distance between corresponding feature vectors:
\begin{equation}
\delta(x_i, y_j) = \|x_i - y_j\|_2
\label{eq:local_distance}
\end{equation}
For univariate signals ($d=1$), this reduces to
$\delta(x_i, y_j) = |x_i - y_j|$.

The DTW cumulative cost matrix $\mathbf{D} \in \mathbb{R}^{n \times m}$
is computed recursively as:
\begin{equation}
\mathbf{D}(i, j) = \delta(x_i, y_j) + \min \begin{cases}
\mathbf{D}(i-1, j) & \text{(vertical/insertion)} \\
\mathbf{D}(i, j-1) & \text{(horizontal/deletion)} \\
\mathbf{D}(i-1, j-1) & \text{(diagonal/match)}
\end{cases}
\label{eq:dtw_cost}
\end{equation}
with boundary conditions:
\begin{align}
\mathbf{D}(1,1) &= \delta(x_1, y_1) \notag \\
\mathbf{D}(i,1) &= \mathbf{D}(i-1,1) + \delta(x_i, y_1) \quad
    \text{for } i > 1 \notag \\
\mathbf{D}(1,j) &= \mathbf{D}(1,j-1) + \delta(x_1, y_j) \quad
    \text{for } j > 1
\end{align}
The final DTW distance between $X$ and $Y$ is
$\text{DTW}(X,Y) = \mathbf{D}(n,m)$.

\subsubsection{Optimal Warping Path}
\label{sec:dtw_path}

% ── NEW FIGURE (replaces PD vs PD single panel) ──────────────────────
\begin{figure*}[t]
\centering
\includegraphics[width=\textwidth]{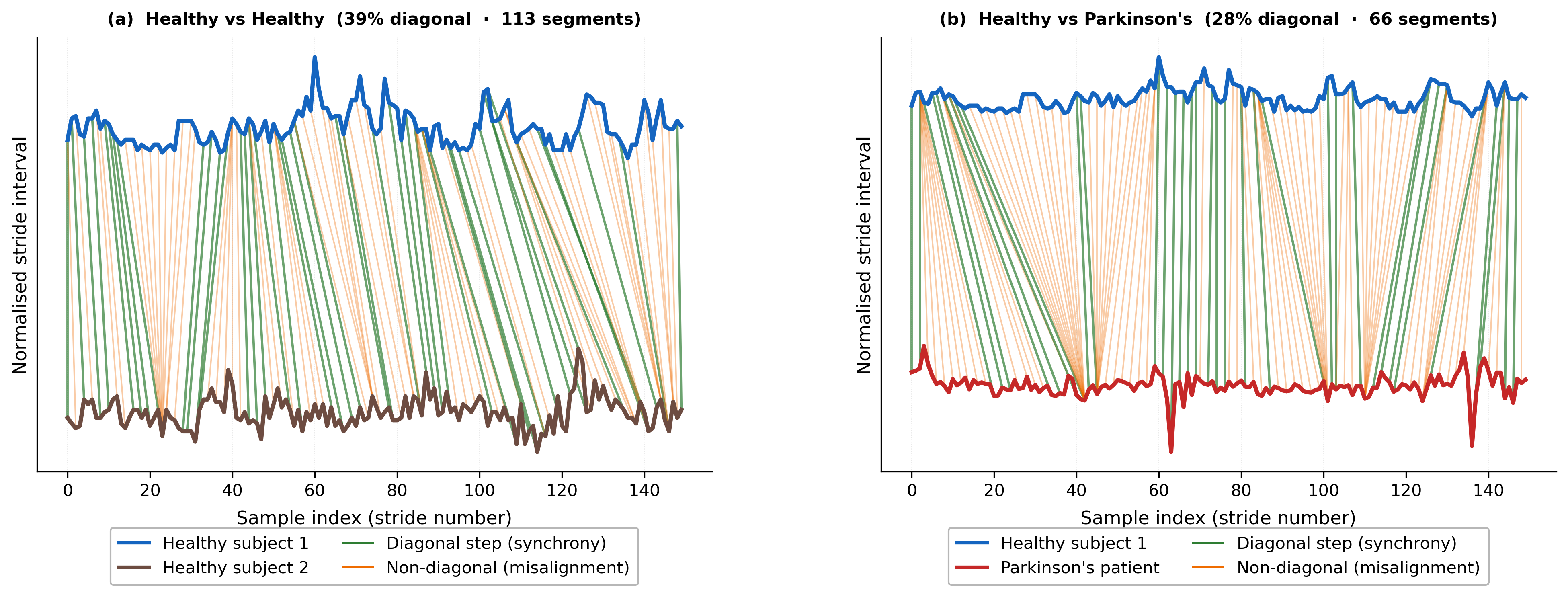}
\caption{DTW point-to-point alignment for two representative
pairs from GaitDB ($n=150$ strides each), demonstrating that
DTW reveals clinically meaningful temporal structure.
Green connectors indicate diagonal steps where both sequences
advance together (sustained temporal synchrony); orange
connectors indicate non-diagonal steps where one sequence
waits (temporal misalignment).
\textbf{(a) Healthy vs Healthy (39\% diagonal, 113 segments):}
two healthy subjects share substantial rhythmic gait structure
recoverable by DTW, confirming that alignment is both meaningful
and informative for these signals.
\textbf{(b) Healthy vs Parkinson's (28\% diagonal, 66 segments):}
alignment with a Parkinson's patient yields fewer diagonal steps
and fewer segments, reflecting stride irregularity and temporal
fragmentation characteristic of pathological gait.
DiaSeg extracts maximal contiguous diagonal runs as geometric
features; the contrast between panels directly motivates why
segment count and length distribution discriminate healthy
from pathological gait.}
\label{fig:dtw_visualization}
\end{figure*}
% ─────────────────────────────────────────────────────────────────────

The optimal warping path $\mathcal{P} = (p_1, p_2, \ldots, p_K)$
where $p_k = (i_k, j_k)$ is obtained by backtracking from $(n,m)$
to $(1,1)$. At each step, we select the predecessor cell with
minimum cumulative cost:
\begin{equation}
p_{k-1} = \argmin_{p \in \mathcal{N}(p_k)} \mathbf{D}(p)
\label{eq:backtrack}
\end{equation}
where $\mathcal{N}(i,j) = \{(i-1,j), (i,j-1), (i-1,j-1)\}$
denotes the set of valid predecessor cells.

Each transition $p_k \to p_{k-1}$ has an associated direction:
\begin{equation}
\text{dir}(p_k, p_{k-1}) = \begin{cases}
\text{diagonal}   & \text{if } \Delta i = 1,\ \Delta j = 1 \\
\text{vertical}   & \text{if } \Delta i = 1,\ \Delta j = 0 \\
\text{horizontal} & \text{if } \Delta i = 0,\ \Delta j = 1
\end{cases}
\label{eq:direction}
\end{equation}
where $\Delta i = i_k - i_{k-1}$ and $\Delta j = j_k - j_{k-1}$.
A diagonal step indicates simultaneous progression through both
sequences, representing temporal synchrony. Vertical and horizontal
steps correspond to temporal insertions and deletions, respectively,
indicating local temporal misalignment.
In practice, the stride interval and accelerometer signals
used in this study are continuously varying, making exact ties in
the cumulative cost matrix measure-zero events that do not arise
empirically, verified across all 30 pairwise comparisons in our
experiments. Figure~\ref{fig:dtw_visualization} illustrates these
step types on two representative pairs from GaitDB, where the
contrast in diagonal step proportions (39\% healthy--healthy vs
28\% healthy--Parkinson's). This demonstrates that DTW alignment
is meaningful for these signals. Furthermore, our population-level
aggregation over thousands of segments provides robustness to any
residual path variations.

\subsection{Diagonal Segment Extraction with Controlled Breaks}
\label{sec:extraction}

Strict diagonal segments, where every step is diagonal, capture
perfect temporal synchrony but fragment under natural signal
variability. We therefore define a diagonal segment with controlled
breaks as a maximal contiguous subsequence
$\mathcal{S} = (p_s, \ldots, p_e)$ of the optimal path satisfying
three conditions. First, the first and last steps are diagonal.
Second, at most $\ell$ non-diagonal steps occur internally. Third,
for a given break count $\ell$, no strictly longer subsequence with
at most $\ell$ internal non-diagonal steps and diagonal endpoints
contains $\mathcal{S}$. Segments extracted for different values of
$\ell$ are therefore independent and may overlap.
The parameter $\ell_{\max}$ controls the strictness of extraction,
where $\ell_{\max} = 0$ enforces perfect synchrony while
$\ell_{\max} > 0$ permits flexible alignment.
Our forward-scan algorithm identifies maximal segments for each
admissible break count up to $\ell_{\max}$ by starting at strict
run boundaries, defined as positions where a diagonal step
immediately follows a non-diagonal step as given in
Eq.~(\ref{eq:boundary}), and extending forward while tracking
cumulative breaks. For a segment spanning indices $[s, e]$ with
$\ell$ breaks, the effective length is $(e - s + 1) - \ell$,
and degenerate segments with $L \leq \ell$ are discarded.
Algorithm~\ref{alg:diagonal_extraction} details the full procedure.
Its worst-case complexity is $O(K^2)$ where $K \approx \max(n, m)$,
matching the DTW computation. In practice, for small $\ell_{\max}$,
the inner loop terminates after encountering $\ell_{\max}+1$
non-diagonal steps, yielding $O(K \cdot \ell_{\max} \cdot g)$
where $g$ is the mean spacing between non-diagonal steps.

\begin{equation}
\text{dir}(p_k, p_{k-1}) = \text{diagonal} \land
\text{dir}(p_{k-1}, p_{k-2}) \neq \text{diagonal}
\label{eq:boundary}
\end{equation}

\begin{algorithm}[t]
\caption{Extract Diagonal Segments with Controlled Breaks}
\label{alg:diagonal_extraction}
\begin{algorithmic}[1]
\Require Optimal path $\mathcal{P} = (p_0, p_1, \ldots, p_K)$,
         maximum breaks $\ell_{\max}$
\Ensure Collection of diagonal segments $\mathcal{D}$
\State $\mathcal{D} \gets \{\ell: [] \text{ for }
    \ell = 0, \ldots, \ell_{\max}\}$
\For{$i \gets 1$ to $K$}
    \If{$i < 3$ \textbf{or}
        $\text{dir}(p_i, p_{i-1}) \neq \text{diagonal}$ \textbf{or}
        $\text{dir}(p_{i-1}, p_{i-2}) = \text{diagonal}$}
        \State \textbf{continue}
    \EndIf
    \State $s \gets i - 1$
    \State $\mathcal{E} \gets \{\}$
    \State $\text{breaks} \gets 0$
    \For{$j \gets i$ to $K$}
        \If{$\text{dir}(p_j, p_{j-1}) = \text{diagonal}$}
            \State $\mathcal{E}[\text{breaks}] \gets j$
        \Else
            \State $\text{breaks} \gets \text{breaks} + 1$
            \If{$\text{breaks} > \ell_{\max}$}
                \State \textbf{break}
            \EndIf
        \EndIf
    \EndFor
    \For{each $(\ell, e) \in \mathcal{E}$}
        \State $L \gets (e - s + 1) - \ell$
        \If{$L \leq \ell$}
            \State \textbf{continue}
        \EndIf
        \State $\mathcal{S} \gets \{s, e, L, \ell\}$
        \State $\mathcal{D}[\ell].\text{append}(\mathcal{S})$
    \EndFor
\EndFor
\State \Return $\mathcal{D}$
\end{algorithmic}
\end{algorithm}

\subsection{Feature Quantification}

Each diagonal segment $\mathcal{S}$ is characterized by a
five-dimensional feature vector (Table~\ref{tab:features})
capturing geometric, temporal, and cost-based alignment
properties, enabling unsupervised pattern discovery without
domain expertise.
\begin{table}[t]
\caption{Diagonal Segment Features}
\label{tab:features}
\begin{tabular}{@{}lll@{}}
\toprule
\textbf{Feature} & \textbf{Definition} & \textbf{Interpretation} \\
\midrule
$L$   & $(e-s+1)-\ell$                    & Sustained alignment extent \\
$\ell$ & Break count                       & Alignment flexibility \\
$d$   & $|\mathbf{D}(p_e)-\mathbf{D}(p_s)|$ & Local alignment quality \\
$t_0$ & $s$                                & Temporal/phase position \\
$p_l$ & $|\mathcal{P}|$                    & Global path context \\
\botrule
\end{tabular}
\end{table}
Longer segments ($L$) with fewer breaks ($\ell$) indicate stronger
temporal correspondence. Low cost variation ($d$) suggests consistent
local similarity. Temporal position ($t_0$) enables phase-specific
pattern identification, while path length ($p_l$) provides
normalization context. For domain-specific applications, metadata
can augment features in gait analysis, including sensor identifier,
limb laterality, gait phase, and cycle identifiers.
The five features were selected to provide a minimal yet
complete geometric characterisation of a diagonal segment,
with one measure of extent ($L$), one of flexibility ($\ell$),
one of alignment quality ($d$), one of temporal localisation
($t_0$), and one of global context ($p_l$). Directionality
features could quantify the angular deviation of a segment
from the strict diagonal, capturing asymmetric temporal
stretching between the two sequences. Curvature features could
measure how much the path bends within a segment, distinguishing
gradual from abrupt temporal shifts. Density features could
capture the local concentration of diagonal steps within a
sliding window, providing a continuous alternative to the
binary break-count representation. Interaction features between
$L$ and $d$ could capture segments that are long but costly,
distinguishing sustained but imperfect synchrony from brief
perfect alignment. We deliberately restricted the feature set
to five dimensions to maintain interpretability and avoid
overfitting on the small clinical cohorts available (12 patients
in BLISS, 15 in GaitDB), where high-dimensional representations
risk fitting noise rather than genuine gait structure.
Systematic evaluation of richer feature sets on larger cohorts
remains an open direction for future work.

\subsection{Unsupervised Clustering Framework}

For $N$ sequences, diagonal segments can be extracted from up to
$\binom{N}{2}$ pairwise DTW alignments. In our experiments,
we use a stratified subset of pairwise comparisons as detailed
in Section~\ref{sec:extraction_results}, forming a feature matrix
$\mathbf{F} \in \mathbb{R}^{M \times p}$ where $M \gg N$ is the
total number of extracted segments and $p$ is the number of
features, with each segment characterized by the five properties
$L$, $\ell$, $d$, $t_0$, and $p_l$ defined in
Section~\ref{sec:extraction}. Each feature dimension $k$ is then
standardized through z-score normalization:
\begin{equation}
\hat{f}_i^{(k)} = \frac{f_i^{(k)} - \mu^{(k)}}{\sigma^{(k)}}
\label{eq:zscore}
\end{equation}

We then apply clustering, using k-means, hierarchical Ward, or GMM,
to the normalized feature matrix to discover segment groups with
similar alignment characteristics. To stratify sequences according
to their alignment behavior, we aggregate each sequence $X_i$
across its segment cluster memberships into a histogram profile
$\mathbf{h}_i$:
\begin{equation}
\mathbf{h}_i = \left[\frac{|\{\mathcal{S} \in \mathcal{C}_j :
X_i \in \mathcal{S}\}|}{|\{\mathcal{S} : X_i \in \mathcal{S}\}|}
\right]_{j=1}^k
\label{eq:sequence_profile}
\end{equation}
This profile enables data-driven discovery of temporal pattern
groups without predefined features, as sequences sharing similar
alignment behaviors will naturally produce similar histogram
profiles and thus cluster together.

% ====================================================================
\section{Experimental Validation}
\label{sec:experiments}
% ====================================================================

We validate our diagonal segment extraction framework through
a systematic multi-stage evaluation. First, we demonstrate
unsupervised pattern discovery on unlabeled neurodegenerative
gait data to establish that segments form meaningful natural
groupings. Second, we validate against ground truth biomechanical
phase annotations to confirm alignment with physiological events.
Third, we perform supervised pathology discrimination using
labeled clinical data to quantify discriminative capacity.
Finally, we compare with traditional cycle-based methods to
position the complementary utility of local versus global features.

\subsection{Datasets and Experimental Setup}
\label{sec:datasets_setup}
This section describes the datasets used for validation
(Section~\ref{sec:datasets_desc}), the evaluation metrics
adopted to assess clustering and classification performance
(Section~\ref{sec:metrics}), and the implementation details
of the DiaSeg framework (Section~\ref{sec:implementation}).

\subsubsection{Datasets Description}
\label{sec:datasets_desc}

We evaluated on three independent gait datasets comprising
91 subjects across multiple clinical conditions
(Table~\ref{tab:datasets}). GaitDB and GaitNDD from
PhysioNet~\cite{goldberger2000physiobank} datasets provide stride interval
time series from neurodegenerative diseases including Parkinson's,
Huntington's, and ALS. These datasets enable unsupervised pattern
discovery without ground truth labels. BLISS dataset \cite{bata1425} provides continuous
acceleration sequences with seven-phase gait annotations spanning
Loading Response through Terminal Swing, enabling both ground truth
biomechanical validation and supervised classification with known
pathology labels comprising 8 healthy controls and 4 pathological
patients.
\begin{table*}[ht]
\centering
\caption{Gait Dataset Characteristics.}
\label{tab:datasets}
\footnotesize
\setlength{\tabcolsep}{4pt}
\begin{tabular}{@{}lclllp{4.2cm}@{}}
\toprule
\textbf{Dataset} & \textbf{N} & \textbf{Sensor}
& \textbf{Setting} & \textbf{Annotations} & \textbf{Populations} \\
\midrule
GaitDB~\cite{hausdorff2000maturation}
  & 15 & Force insoles & Overground
  & Stride intervals
  & 5 young healthy, 5 elderly, 5 Parkinson's \\[0.5ex]
GaitNDD~\cite{hausdorff2018gait}
  & 64 & Force insoles & Overground
  & Stride intervals
  & 16 healthy, 15 Parkinson's, 20 Huntington's, 13 ALS \\[0.5ex]
BLISS~\cite{bata1425}
  & 12 & Accelerometer & Treadmill
  & \textbf{7 phase labels}
  & 8 healthy, 2 mild, 2 severe (brain tumor, stroke) \\[0.5ex]
\midrule
\textbf{Total} & \textbf{91} & -- & -- & --
  & 6 conditions across 3 independent datasets \\
\botrule
\end{tabular}
\end{table*}
All sequences underwent z-score standardization before DTW
computation to ensure temporal features contribute equally
regardless of individual baseline walking speed differences.

\subsubsection{Evaluation Metrics}
\label{sec:metrics}

We assess clustering quality and classification performance using
complementary unsupervised and supervised metrics. For unsupervised
evaluation, the Silhouette Score~\cite{shahapure2020cluster}
measures sample-cluster fit ranging from $-1$ to $+1$ where scores
exceeding $0.25$ indicate meaningful structure for biological
data~\cite{ogbuabor2018clustering}. The Davies-Bouldin
Index~\cite{xiao2017davies} evaluates cluster separation through
the ratio of within-cluster scatter to centroid distance, with
lower values indicating better separation. For supervised
evaluation, we report accuracy, precision, recall, and F1-score
computed via stratified cross-validation. Statistical significance
is assessed through two-tailed t-tests with significance threshold
$p < 0.05$. In addition, we report Adjusted Rand Index (ARI),Normalised Mutual Information (NMI), Purity, and V-measure
computed against the available clinical labels, providing
external validation of the discovered cluster structure
against known ground truth.

\subsubsection{Implementation Details}
\label{sec:implementation}

We implemented the framework in Python 3.9 using scikit-learn 1.3
and NumPy 1.24. The maximum break tolerance $\ell_{\max}$ is set
adaptively per pairwise comparison as a proportion of the sequence
length, rather than as a fixed global value. In practice, this
yielded values not exceeding 3 across all sequences in the
evaluated datasets, reflecting the relatively uniform temporal
extent of the gait recordings used. Sequences of greater length
or higher temporal complexity may naturally admit larger
$\ell_{\max}$ values under this scheme.
Table~\ref{tab:lmax_sensitivity} confirms that $\ell_{\max}=3$
represents the point of diminishing returns: segment count
increases 120\% from $\ell_{\max}=0$ to $\ell_{\max}=3$ while
silhouette remains above 0.25, with only 14\% additional segments
gained by increasing to $\ell_{\max}=4$, validating the adaptive
selection scheme.
All features underwent z-score normalization before clustering
and classification. For patient-level aggregation, RobustScaler
was applied instead of StandardScaler, as the small 12-patient
cohort makes z-score normalization sensitive to individual outlier
patients; RobustScaler's median and IQR-based scaling provides
more stable normalization under these conditions.
For unsupervised validation, we evaluated K-means, hierarchical
Ward, and GMM algorithms. For supervised classification, we
employed SVM with RBF kernel (parameters $C=1.0$, $\gamma=$scale)
and Logistic Regression with L2 regularization ($C=1.0$).

\subsection{Diagonal Segment Extraction Results}
\label{sec:extraction_results}
\begin{figure*}[ht]
\centering
\includegraphics[width=0.95\textwidth]{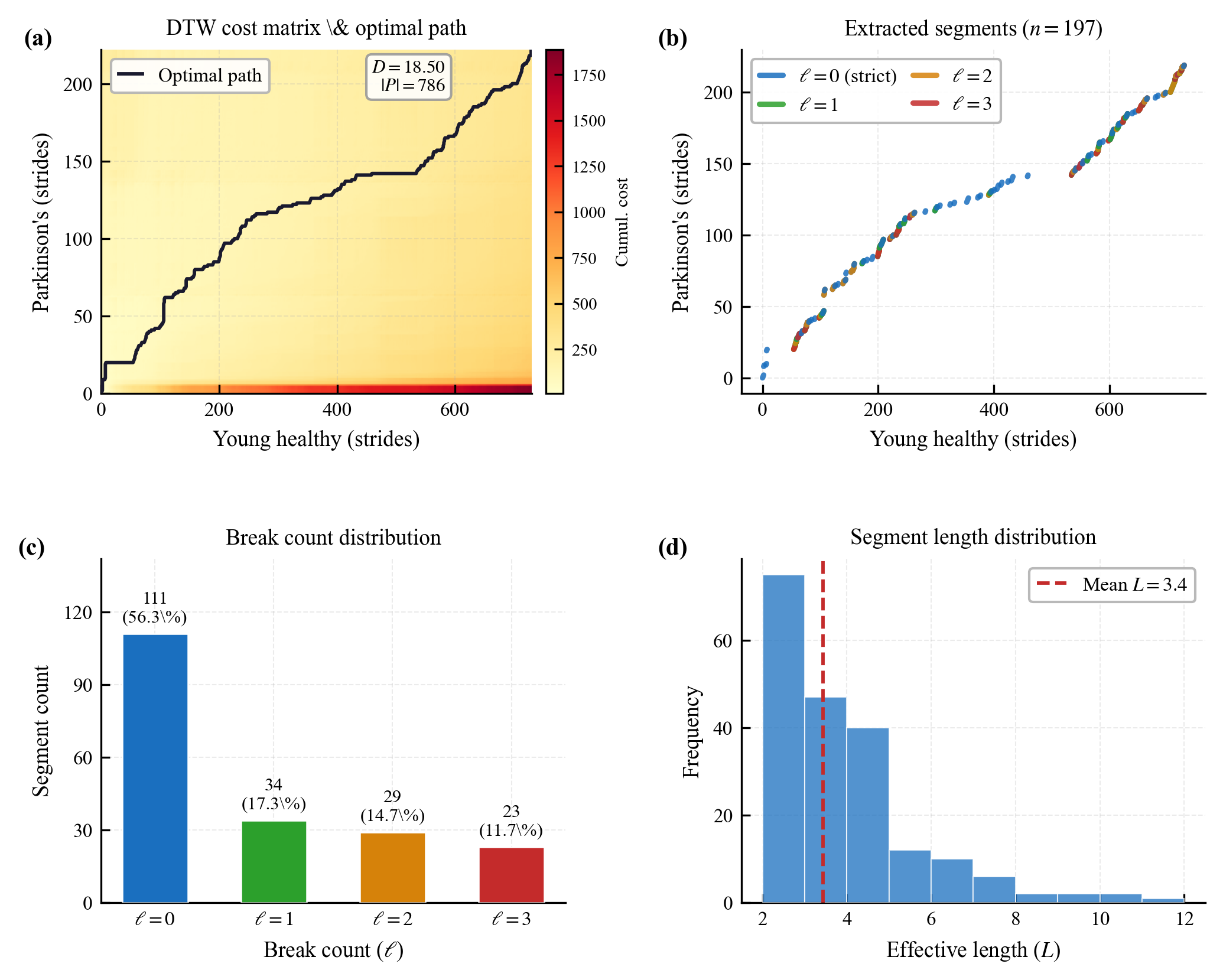}
\caption{Diagonal segment extraction from a representative DTW
optimal path (young healthy vs.\ Parkinson's patient from GaitDB).
\textbf{(a)}~Cost matrix with optimal path (blue); color intensity
indicates accumulated cost. \textbf{(b)}~Extracted segments
($n=197$) color-coded by break count: $\ell=0$ (blue), $\ell=1$
(green), $\ell=2$ (orange), $\ell=3$ (red).
\textbf{(c)}~Break distribution showing strict segments as the
dominant category.
\textbf{(d)}~Length distribution spanning 2--11 strides
(mean 3.4).}
\label{fig:extraction_overview}
\end{figure*}
We extracted diagonal segments from 30 pairwise comparisons,
15 from GaitDB yielding 5,602 segments and 15 from GaitNDD
yielding 1,610 segments, for a total of 7,212 segments across
all pairwise alignments. Figure~\ref{fig:extraction_overview}
illustrates the process for a representative healthy versus
Parkinson's comparison from GaitDB, from which 197 segments
were recovered and color-coded by break count. Strict segments
($\ell=0$) dominate at 56.3\%, with progressively fewer
segments at $\ell=1$ (17.3\%), $\ell=2$ (14.7\%), and
$\ell=3$ (11.7\%), and segment lengths span 2--11 strides
with a mean of 3.4, reflecting both brief synchronization
episodes and extended coordination patterns.
Table~\ref{tab:segment_extraction} stratifies results by
comparison type across both datasets. Within-healthy
comparisons yield the highest segment counts and longest
mean lengths (3.8--4.5 strides), reflecting stable rhythmic
gait, while within-pathology comparisons produce the fewest
segments, with only 13 in GaitDB and 45 in GaitNDD, as
individual compensatory strategies fragment temporal
correspondence even between identically diagnosed patients.
Cross-condition comparisons fall between these extremes,
suggesting that some alignment capacity persists despite
pathological disruption. Together, segment count and mean
length serve as sensitive markers of temporal coordination
capacity.
Table~\ref{tab:lmax_sensitivity} confirms that $\ell_{\max}=3$
represents the point of diminishing returns: segment count
increases 120\% from $\ell_{\max}=0$ to $\ell_{\max}=3$
while silhouette remains above 0.25 across all tested values,
with only 14\% additional segments gained by increasing to
$\ell_{\max}=4$, justifying the adaptive selection of
$\ell_{\max} \leq 3$ across all evaluated sequences.

\begin{table}[t]
\caption{Diagonal Segment Extraction Statistics by Comparison
Type. The stratified subset shown (5 pairs per comparison type,
per dataset) yields 1,852 segments; full extraction across all
30 pairwise alignments yields 7,212 segments total used in
downstream clustering.}
\label{tab:segment_extraction}
\begin{tabular}{@{}llrr@{}}
\toprule
\textbf{Dataset} & \textbf{Comparison Type}
& \textbf{Segments} & \textbf{Mean $L$} \\
\midrule
\multirow{3}{*}{GaitDB}
  & Healthy--Healthy     & 462            & 3.8 \\
  & Cross-condition      & 616            & 3.6 \\
  & Pathology--Pathology &  13            & 3.5 \\
\cmidrule(lr){2-4}
  & \textit{Subtotal}    & \textit{1,091} & \textit{3.6} \\
\midrule
\multirow{3}{*}{GaitNDD}
  & Healthy--Healthy     & 156            & 4.5 \\
  & Cross-condition      & 560            & 3.8 \\
  & Pathology--Pathology &  45            & 3.7 \\
\cmidrule(lr){2-4}
  & \textit{Subtotal}    & \textit{761}   & \textit{3.8} \\
\midrule
\multicolumn{2}{@{}l}{\textbf{Stratified Total}}
  & \textbf{1,852} & \textbf{3.8} \\
\botrule
\end{tabular}
\end{table}

\begin{table}[t]
\caption{Effect of break tolerance $\ell_{\max}$ on segment
count and clustering quality on GaitDB (K-means, $k=2$),
computed on the same 15 stratified pairs as
Table~\ref{tab:clustering}. Performance remains stable
across $\ell_{\max} \in \{0,1,2,3,4\}$, with silhouette
exceeding the 0.25 meaningful structure threshold throughout.
\textbf{Bold} indicates the value selected as the point
of diminishing returns.}
\label{tab:lmax_sensitivity}
\begin{tabular}{@{}cccc@{}}
\toprule
$\ell_{\max}$ & Segments
& Silhouette $\uparrow$ & DB Index $\downarrow$ \\
\midrule
0 & 3{,}423 & 0.401 & 1.267 \\
1 & 4{,}994 & 0.341 & 1.378 \\
2 & 6{,}347 & 0.310 & 1.390 \\
\textbf{3} & \textbf{7{,}549} & \textbf{0.318} & \textbf{1.336} \\
4 & 8{,}635 & 0.322 & 1.297 \\
\botrule
\end{tabular}
\footnotetext{Silhouette computed on z-scored features
(StandardScaler) on the same 15 stratified pairs used
throughout. Segment count increases 120\% from
$\ell_{\max}=0$ to $\ell_{\max}=3$; marginal gain to
$\ell_{\max}=4$ is only 14\%, justifying $\ell_{\max}=3$.}
\end{table}

% ── UPDATED TABLE: BLISS added ────────────────────────────────────────
\begin{table}[t]
\caption{Unsupervised Clustering Performance on Diagonal Segments.
Bold indicates best result per dataset. DB Index: lower is better;
Silhouette: higher is better.}
\label{tab:clustering}
\begin{tabular}{@{}llcc@{}}
\toprule
\textbf{Dataset} & \textbf{Method ($k$)}
& \textbf{Silhouette $\uparrow$} & \textbf{DB Index $\downarrow$} \\
\midrule
\multirow{4}{*}{GaitDB}
  & K-means ($k=2$)      & \textbf{0.337} & \textbf{1.257} \\
  & K-means ($k=3$)      & 0.253          & 1.280          \\
  & Hierarchical ($k=2$) & 0.260          & 1.505          \\
  & GMM ($k=2$)          & 0.265          & 1.408          \\
\midrule
\multirow{4}{*}{GaitNDD}
  & K-means ($k=2$)      & \textbf{0.329} & \textbf{1.272} \\
  & K-means ($k=3$)      & 0.237          & 1.386          \\
  & Hierarchical ($k=2$) & 0.288          & 1.405          \\
  & GMM ($k=2$)          & 0.249          & 1.472          \\
\midrule
\multirow{4}{*}{BLISS}
  & K-means ($k=2$)      & 0.398          & 1.183 \\
  & K-means ($k=3$)      & \textbf{0.411} & \textbf{1.084} \\
  & Hierarchical ($k=2$) & 0.308          & 1.421 \\
  & GMM ($k=2$)          & 0.311         & 1.478 \\
\botrule
\end{tabular}
\end{table}
% ── NEW TABLE: External metrics ───────────────────────────────────────
\begin{table}[t]
\caption{Clustering quality metrics against clinical labels
(segment-level). ARI: Adjusted Rand Index; NMI: Normalised
Mutual Information. Binary setting: healthy vs pathological;
3-class GaitDB: young vs elderly vs Parkinson's.
Bold indicates best result per dataset and setting.}
\label{tab:external_metrics}
\begin{tabular}{@{}lllcccc@{}}
\toprule
\textbf{Dataset} & \textbf{Setting} & \textbf{Method}
& \textbf{ARI} & \textbf{NMI}
& \textbf{Purity} & \textbf{V} \\
\midrule
\multirow{6}{*}{GaitDB}
  & \multirow{3}{*}{Binary ($k=2$)}
    & K-means & 0.966 & 0.921 & 0.995 & 0.921 \\
  & & Ward    & \textbf{0.986} & \textbf{0.963} & \textbf{0.998} & \textbf{0.963} \\
  & & GMM     & 0.404 & 0.364 & 0.895 & 0.364 \\
\cmidrule(lr){2-7}
  & \multirow{3}{*}{3-class ($k=3$)}
    & K-means & 0.193 & 0.355 & \textbf{0.733} & 0.355 \\
  & & Ward    & 0.196 & 0.360 & 0.734 & 0.360 \\
  & & GMM     & \textbf{0.234} & \textbf{0.360} & 0.730 & \textbf{0.360} \\
\midrule
\multirow{3}{*}{BLISS}
  & \multirow{3}{*}{Binary ($k=2$)}
    & K-means & 0.194 & 0.072 & \textbf{0.801} & 0.072 \\
  & & Ward    & 0.169 & 0.065 & \textbf{0.801} & 0.065 \\
  & & GMM     & \textbf{0.223} & \textbf{0.100} & \textbf{0.801} & \textbf{0.100} \\
\botrule
\end{tabular}
\end{table}

\subsection{Unsupervised Pattern Discovery}
\label{sec:unsupervised_clustering}

\begin{figure*}[ht]
\centering
\includegraphics[width=0.95\textwidth]{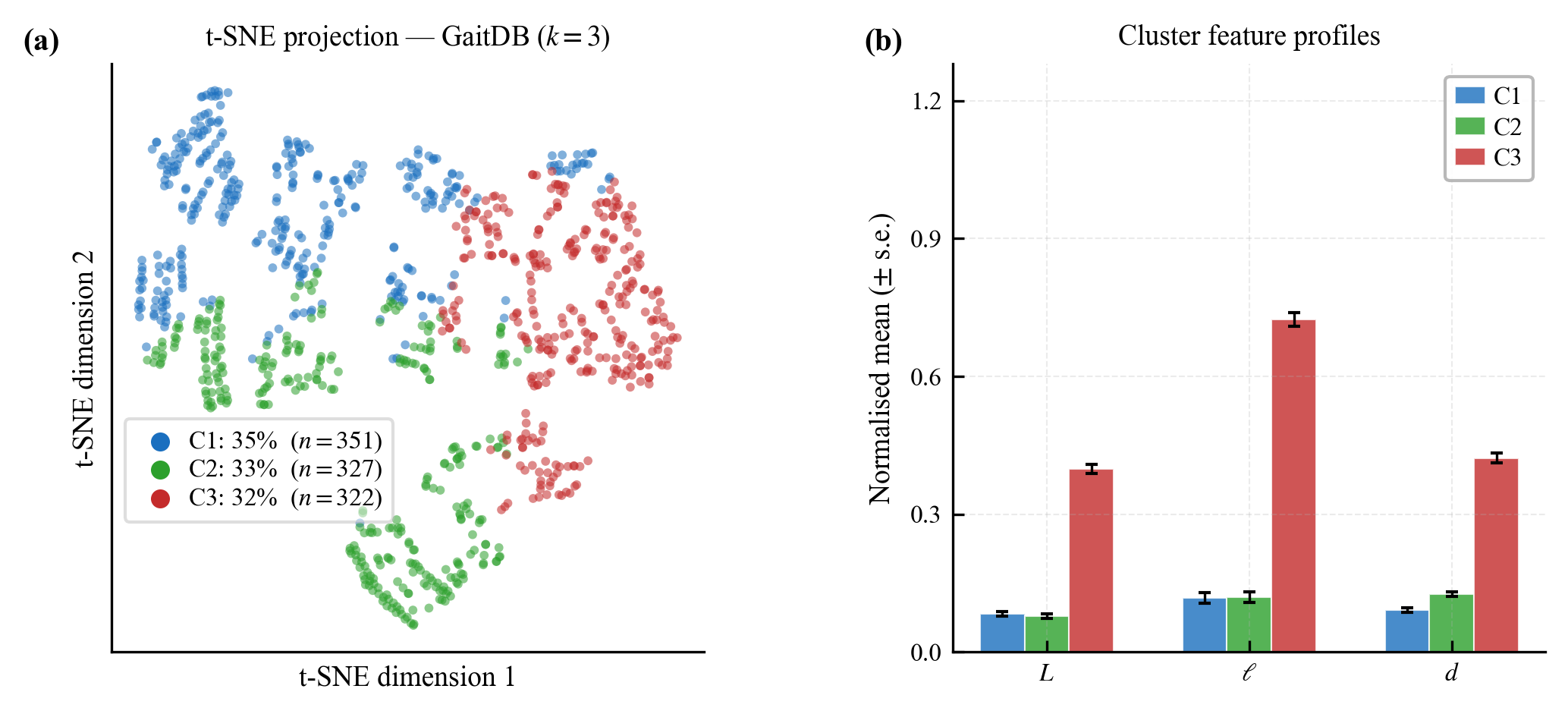}
\caption{t-SNE projection of diagonal segments extracted from
GaitDB, visualised with K-means ($k=3$) to reveal interpretable
alignment subtypes beyond the statistically optimal $k=2$
(see Table~\ref{tab:clustering}).
\textbf{(a)}~Three spatially separated regions: C1 (blue, 35\%,
$n=351$), C2 (green, 33\%, $n=327$), and C3 (red, 32\%, $n=322$).
\textbf{(b)}~Normalised mean feature profiles per cluster
($\pm$~s.e.), showing that C3 exhibits substantially higher
$\ell$ and $d$, identifying it as the flexible high-cost
alignment cluster, while C1 and C2 show similar profiles on
$L$, $\ell$, and $d$ with their separation driven primarily
by temporal position $t_0$ and path context $p_l$.}
\label{fig:tsne_clustering}
\end{figure*}

To validate that diagonal segments form meaningful patterns without
supervision, we performed clustering on the 5-dimensional feature
space comprising segment length $L$, break count $\ell$, DTW
distance $d$, temporal position $t_0$, and path length $p_l$.
Crucially, while prior work applies K-means directly to scalar DTW
distances or hand-crafted biomechanical features, here K-means
operates on structural path features derived from the warping path
geometry itself, making the clustering algorithm a tool for
organizing alignment structure rather than a substitute for it.
We evaluated K-means, hierarchical Ward, and GMM algorithms to
assess whether natural groupings emerge from segment geometry and
alignment properties alone. Table~\ref{tab:clustering} summarizes
clustering performance across methods and datasets.
K-means with $k=2$ achieves the best quantitative performance
across both datasets (silhouette 0.337 for GaitDB, 0.329 for
GaitNDD), exceeding the 0.25 threshold for meaningful biological
structure~\cite{ogbuabor2018clustering}. Davies-Bouldin indices
of 1.257--1.272 confirm reasonable cluster separation.
Hierarchical Ward and GMM methods yield comparable but slightly
lower scores, with consistency across all three algorithms
confirming genuine cluster structure rather than method-specific
artifacts. The optimal $k=2$ for GaitNDD, despite the dataset containing three diagnostic groups (healthy, Parkinson's,
Huntington's, and ALS), reflects a meaningful biological
organisation where segment features primarily capture the
distinction between normal and pathological gait rhythm rather
than disease-specific signatures. This is consistent with the
finding that pathology manifests through distributional shifts
in segment length statistics shared across neurodegenerative
conditions rather than through condition-specific geometric
patterns. Quantitative validation against clinical labels
confirms this interpretation, with binary
healthy--pathological separation achieving ARI up to 0.986
while three-class separation yields ARI of 0.193--0.234
(Table~\ref{tab:external_metrics}), indicating that the
healthy--pathological boundary is strongly captured at the
segment level whereas individual disease boundaries are not.
For BLISS, the optimal cluster count is $k=3$
(silhouette 0.411), consistent with the three severity levels
present in the dataset, namely healthy controls, mild pathology
(brain tumor, stroke with moderate impairment), and severe
pathology (brain tumor, stroke with significant motor
disruption), suggesting that segment features capture disease
severity gradients beyond binary healthy--pathological
classification.
The moderate silhouette scores are consistent with the continuous,
overlapping nature of biological variability in gait alignment.
Neurological disorders introduce individual compensatory strategies
that produce gradual rather than discrete distributional shifts,
and the feature space spanning continuous ranges for length
(2--12 strides) and breaks (0--3) contains no natural
discontinuities that would produce well-separated clusters.
The convergence of structure across K-means, hierarchical Ward,
and GMM is therefore the stronger evidence of genuine organisation.
To examine finer-grained alignment structure beyond the optimal
$k=2$, Figure~\ref{fig:tsne_clustering} visualises $k=3$ clusters
on GaitDB, revealing three interpretable pattern types without
substantially degrading cluster quality (silhouette 0.253,
Table~\ref{tab:clustering}).
Figure~\ref{fig:tsne_clustering}(a) shows three spatially
separated regions in the t-SNE projection, confirming that the
additional cluster captures genuine structure rather than arbitrary
subdivision. Figure~\ref{fig:tsne_clustering}(b) identifies what
distinguishes each cluster through normalised feature profiles.
C3 is clearly distinguished from C1 and C2 by substantially
higher values across all three primary features, namely
effective length $L$, break count $\ell$, and cost variation
$d$, characterising it as the flexible high-cost alignment
cluster corresponding to long segments with frequent breaks
and high alignment cost. C1 and C2 show similar profiles on
$L$, $\ell$, and $d$, with their spatial separation in the
t-SNE projection driven primarily by temporal position $t_0$
and path context $p_l$, suggesting they represent early-cycle
and late-cycle alignment patterns respectively. This three-cluster
interpretation provides richer clinical insight than the binary
$k=2$ partition, at the cost of a modest reduction in
silhouette score.

\subsection{Biomechanical Ground Truth Validation}
\label{sec:ground_truth}

\begin{figure*}[ht]
\centering
\includegraphics[width=0.95\textwidth]{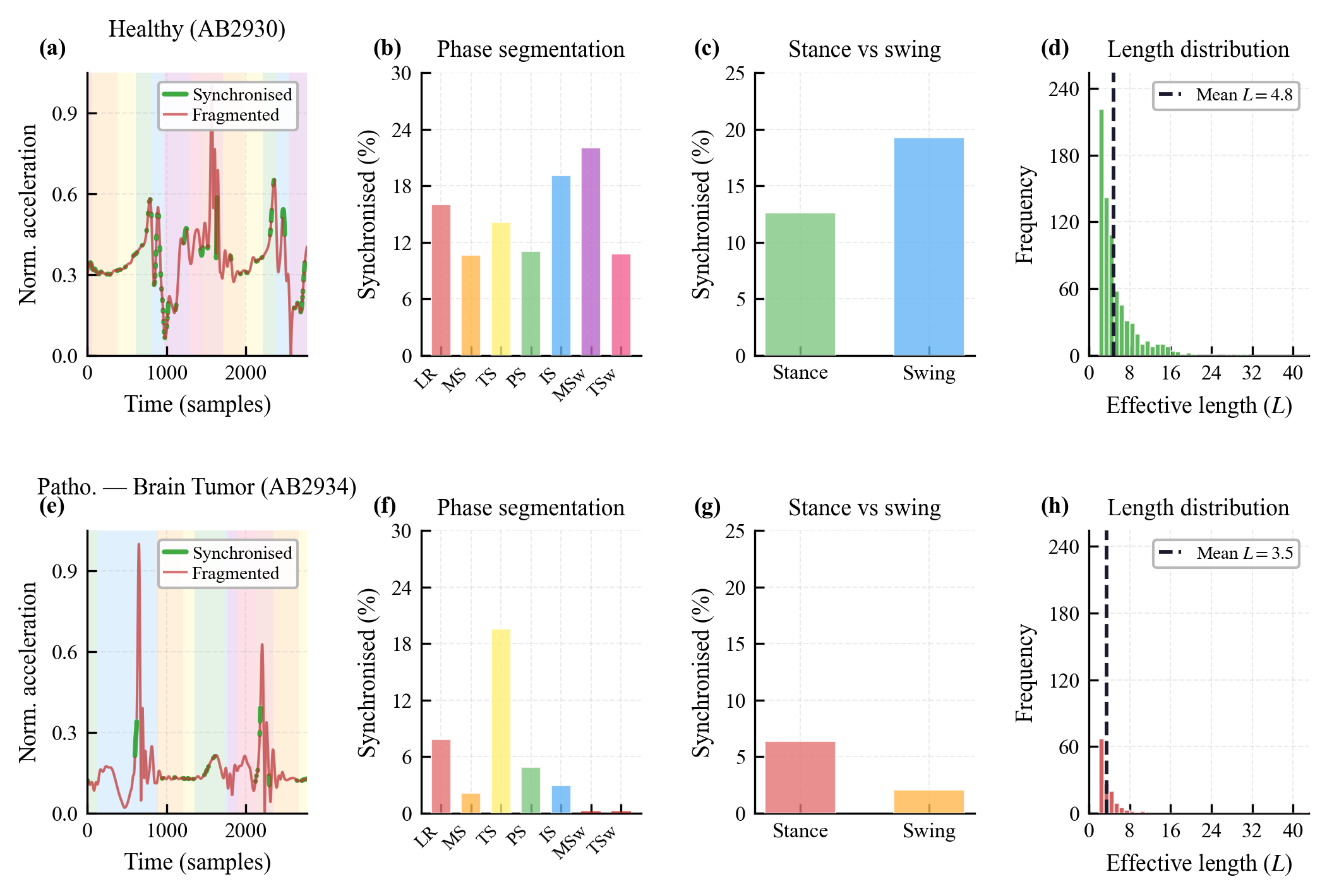}
\caption{Ground truth validation on BLISS dataset (AB2930 healthy
vs.\ AB2934 brain tumor). \textbf{Row~1 --- Healthy control:}
\textbf{(a)}~Acceleration signal with gait phase annotations
(coloured backgrounds) and synchronisation overlay
(green = synchronised, red = fragmented); \textbf{(b)}~phase-specific
synchronisation rates across all seven phases; \textbf{(c)}~stance
vs.\ swing aggregation; \textbf{(d)}~segment length distribution
(mean $L = 4.8$). \textbf{Row~2 --- Brain tumor patient:}
\textbf{(e--h)}~corresponding subfigures showing global fragmentation,
severely reduced phase synchronisation, degraded stance and swing
rates, and concentration of short segments (mean $L = 3.5$).
Identical $y$-axis scales across paired subfigures enable direct
visual comparison.}
\label{fig:bliss_validation}
\end{figure*}

To illustrate how diagonal segment features relate to known
biomechanical events, we present a qualitative case study
comparing one healthy control (AB2930) with one brain tumor
patient (AB2934) from the BLISS dataset, which provides
continuous gait phase annotations spanning seven
biomechanically defined phases such as Loading Response (LR),
Mid Stance (MS), Terminal Stance (TS), Pre-Swing (PS), Initial
Swing (IS), Mid Swing (MSw), and Terminal Swing (TSw). This
comparison serves as an interpretive demonstration of the
framework's ability to localize coordination breakdowns within
the gait cycle rather than as a statistical claim across the
full cohort. Findings should therefore be read as
hypothesis-generating observations pending replication in
larger samples.
Figure~\ref{fig:bliss_validation} examines both subjects
across four complementary analytical perspectives.
The healthy control (row~1) displays extensive synchronised
segments distributed across all seven gait phases, with green
overlay dominating Figure~\ref{fig:bliss_validation}(a). The brain tumor patient
(row~2) shows predominantly fragmented regions with only
sporadic synchronisation visible in Figure~\ref{fig:bliss_validation}(e), consistent
with widespread motor disruption rather than a phase-specific
deficit, though this interpretation is based on a single
pathological case and cannot be generalised without further
evidence.
Phase-specific quantification (Figures~\ref{fig:bliss_validation}(b) and~\ref{fig:bliss_validation}(f)) further
illustrates this pattern. The healthy control maintains
synchronisation rates of 10--23\% across stance phases
(LR, MS, TS, PS) and 19--22\% across swing phases (IS, MSw,
TSw). The pathological case shows reduced synchronisation
across all seven phases, with swing phases approaching zero.
Red-outlined bars in Figure~\ref{fig:bliss_validation}(f) identify phases whose
synchronisation falls below 30\% of the pathological mean,
illustrating how the framework could highlight candidate phases
of severe local disruption for follow-up investigation.
Stance versus swing aggregation (Figures~\ref{fig:bliss_validation}(c) and~\ref{fig:bliss_validation}(g)) reinforces
this observation. The healthy control achieves 12\% stance and
19\% swing synchronisation, while the pathological case shows
markedly lower values in both domains (approximately 6\% stance,
2\% swing), suggesting impairment across the full gait cycle.
Segment length distributions (Figures~\ref{fig:bliss_validation}(d) and~\ref{fig:bliss_validation}(h)) reveal an
observable difference in sustained coordination capacity between
the two subjects. The healthy control exhibits a broad
right-tailed distribution with mean $L = 4.8$ strides and a
long tail extending beyond 32 strides, while the brain tumor
patient shows concentration at short lengths with mean $L = 3.5$
and negligible counts beyond 8 strides. The combination of
reduced mean length and collapsed distributional spread
represents a quantitative pattern of potential interest as a
severity indicator, though validation across larger and more
diagnostically diverse cohorts is necessary before any clinical
interpretation can be drawn.

\subsection{Supervised Pathology Discrimination}
\label{sec:supervised_classification}

In this section, we assess the discriminative capacity of diagonal
segment features by performing supervised and unsupervised
classification on the BLISS dataset \cite{bata1425}, which provides known
pathology labels for 8 healthy controls and 4 pathological cases
grouped into binary classes for analysis.

\subsubsection{Segment-Level Classification}

We extracted 86,917 diagonal segments across all 12 patients,
with individual contributions ranging from 4,468 to 10,834
segments. Class balancing through random undersampling yielded
53,264 segments equally distributed across healthy (26,632)
and pathological (26,632) classes. The feature representation
comprised the five base segment features ($L$, $\ell$, $d$,
$t_0$, $p_l$) defined in Section~\ref{sec:methodology},
with all features z-score normalized prior to classification.
We evaluated classifiers using stratified 5-fold
cross-validation to assess generalization performance as shown in Table~\ref{tab:supervised_segments}. SVM with RBF kernel
achieved 69.3\% $\pm$ 0.3\% accuracy across folds ranging
68.8--69.7\%, while Logistic Regression achieved 69.1\%
$\pm$ 0.4\%. Both results significantly exceed chance level
(50\%) as confirmed by two-tailed t-test ($t(4) = 65.2$,
$p < 0.001$). The minimal variance ($\pm$0.3--0.4\%)
demonstrates reproducibility and confirms genuine discriminative
capacity rather than overfitting to particular data partitions.

\begin{table*}[ht]
\centering
\caption{Segment-Level Classification Results on BLISS Dataset
(5-fold stratified CV).}
\label{tab:supervised_segments}
\begin{tabular}{@{}lcccc@{}}
\toprule
\textbf{Classifier} & \textbf{Accuracy} & \textbf{Precision}
& \textbf{Recall} & \textbf{F1} \\
\midrule
SVM (RBF)
  & 69.3 $\pm$ 0.3 & 69.5 $\pm$ 0.4
  & 69.1 $\pm$ 0.5 & 69.3 $\pm$ 0.3 \\
Logistic Regression
  & 69.1 $\pm$ 0.4 & 69.2 $\pm$ 0.5
  & 69.0 $\pm$ 0.4 & 69.1 $\pm$ 0.4 \\
\midrule
Chance level & 50.0 & 50.0 & 50.0 & 50.0 \\
\botrule
\multicolumn{5}{@{}l}{\footnotesize All results significantly
above chance ($p < 0.001$, two-tailed $t$-test).}
\end{tabular}
\end{table*}

\subsubsection{Patient-Level Aggregation}

To assess whether segment distributions differ systematically
at the patient level, we computed comprehensive aggregate
statistics characterizing the complete segment population for
each patient. For each of the three primary segment properties,
namely length $L$, break count $\ell$, and DTW distance $d$,
we calculated 15 statistical descriptors comprising central
tendency (mean, median), spread (standard deviation,
interquartile range, minimum, maximum, range), distributional
shape (skewness, kurtosis), and additional derived measures
including coefficient of variation, ratio statistics, and
count-based features, yielding 45 aggregate features
per patient capturing the complete distributional characteristics
of their segment populations. 
Unsupervised K-means clustering with two clusters on these
45-dimensional aggregate features achieved 75.0\% accuracy
(9 of 12 patients correctly grouped) with a silhouette score
of 0.615, indicating well-separated natural groupings.
Notably, this analysis used no pathology labels during
clustering, as the algorithm discovered groupings based purely
on distributional properties which were subsequently evaluated
against known clinical status. The method correctly identified
all 8 healthy patients without exception, demonstrating perfect
specificity for normal gait patterns. Among pathological cases,
the clustering correctly identified 1 of 2 severe patients,
with misclassifications concentrated in the mild pathology
cases whose subtle disruptions may not yet manifest in aggregate
segment statistics.
Supervised SVM classification using Leave-One-Out
cross-validation, appropriate for the small 12-patient sample,
achieved 66.7\% accuracy (8 of 12 correct). On this cohort,
unsupervised clustering (75.0\%) and supervised SVM LOO (66.7\%)
differ by one patient (8.3\%), a margin too small to support a
general claim of superiority. Both results are reported as
exploratory estimates pending validation on larger cohorts.
Table~\ref{tab:patient_results} details classification outcomes
for each patient under both approaches, revealing that
performance differences stem primarily from handling of
borderline mild pathology cases. The superior patient-level
performance (75.0\%) compared to segment-level classification
(69.3\%) demonstrates that pathology manifests through aggregate
distributional patterns rather than properties of individual
segments, consistent with clinical understanding that
neurological disorders affect gait variability and coordination
consistency across multiple cycles rather than producing
consistently abnormal individual movement patterns.

\begin{table}[ht]
\centering
\caption{Patient-Level Classification Results.
{\color{green!60!black}\cmark}~= correct,
{\color{red!70!black}\xmark}~= incorrect.}
\label{tab:patient_results}
\begin{tabular}{@{}llcc@{}}
\toprule
\textbf{Patient} & \textbf{Severity}
& \textbf{K-means} & \textbf{SVM LOO} \\
\midrule
AB2930 & Healthy
  & {\color{green!60!black}\cmark}
  & {\color{green!60!black}\cmark} \\
AB2931 & Healthy
  & {\color{green!60!black}\cmark}
  & {\color{red!70!black}\xmark} \\
AB2933 & Healthy
  & {\color{green!60!black}\cmark}
  & {\color{red!70!black}\xmark} \\
AB2937 & Healthy
  & {\color{green!60!black}\cmark}
  & {\color{green!60!black}\cmark} \\
AB2938 & Healthy
  & {\color{green!60!black}\cmark}
  & {\color{green!60!black}\cmark} \\
AB2939 & Healthy
  & {\color{green!60!black}\cmark}
  & {\color{green!60!black}\cmark} \\
AB2940 & Healthy
  & {\color{green!60!black}\cmark}
  & {\color{green!60!black}\cmark} \\
AB2941 & Healthy
  & {\color{green!60!black}\cmark}
  & {\color{red!70!black}\xmark} \\
\midrule
AB2932 & Mild
  & {\color{red!70!black}\xmark}
  & {\color{red!70!black}\xmark} \\
AB2935 & Mild
  & {\color{red!70!black}\xmark}
  & {\color{green!60!black}\cmark} \\
\midrule
AB2934 & Severe
  & {\color{green!60!black}\cmark}
  & {\color{green!60!black}\cmark} \\
AB2936 & Severe
  & {\color{red!70!black}\xmark}
  & {\color{green!60!black}\cmark} \\
\midrule
\textbf{Accuracy} &
  & \textbf{9/12 (75.0\%)}
  & \textbf{8/12 (66.7\%)} \\
\botrule
\end{tabular}
\end{table}

\subsubsection{Feature Importance Analysis}

Feature importance analysis reveals scale-dependent discriminative
mechanisms explaining the performance difference between
segment-level (69\%) and patient-level (75\%) classification.
At the segment level, DTW distance to the healthy reference
dominates discrimination, while individual segment geometry
features ($L$, $\ell$) contribute minimally, suggesting that
at the individual segment scale, overall alignment quality
matters more than the geometric properties of specific segments.
At the patient level, PCA of the 45 aggregate features
reveals that the first principal component explains 70.7\% of
variance and is most strongly loaded by distributional shape
features of segment length, with $L$ kurtosis (loading 0.796),
$L$ skewness (0.344), and $L$ range (0.209) showing the highest
PC1 loadings among all 45 features. This indicates that
pathology manifests primarily through altered statistical
distributions of segment lengths across multiple cycles,
specifically through increased tail weight (kurtosis) and
asymmetry (skewness) of the length distribution, rather
than through mean values or properties of individual segments.
This signature is accessible only through segment extraction
and population-level aggregation.
This scale-dependent inversion validates that diagonal segments
provide discriminative information beyond DTW distance alone.
At the segment level, overall alignment quality (DTW distance)
dominates; at the patient level, the shape of the segment
length distribution emerges as the primary discriminator,
capturing coordination deterioration that individual segment
analysis cannot detect.

\subsection{Complementarity with Cycle-Based Features}
\label{sec:baseline_comparison}
\begin{table}[t]
\caption{Feature Representation Comparison.}
\label{tab:feature_comparison}
\begin{tabular}{@{}lcc@{}}
\toprule
\textbf{Approach} & \textbf{Classifier} & \textbf{Accuracy} \\
\midrule
Diagonal Segments (local)       & SVM            & 69.3\% $\pm$ 0.3\% \\
Patient Aggregation (distributional) & K-means   & 75.0\% \\
\midrule
\multirow{3}{*}{Gait Cycles (global)}
  & Random Forest     & \textbf{91.3\%} $\pm$ 4.2\% \\
  & SVM               & 88.9\% $\pm$ 4.1\% \\
  & Gradient Boosting & 90.1\% $\pm$ 4.7\% \\
\botrule
\end{tabular}
\end{table}

% ── NEW TABLE: Complementarity ────────────────────────────────────────
\begin{table}[t]
\caption{Complementarity of local segment features and global
cycle-level features on BLISS ($n=12$ patients, LOO-CV,
binary: healthy vs pathological).}
\label{tab:complementarity}
\begin{tabular}{@{}llcc@{}}
\toprule
\textbf{Feature Set} & \textbf{Classifier}
& \textbf{Accuracy} & \textbf{F1} \\
\midrule
\multirow{3}{*}{Local (segment)}
  & SVM (RBF) & \textbf{0.917} & \textbf{0.913} \\
  & LR        & \textbf{0.917} & \textbf{0.913} \\
  & RF        & \textbf{0.917} & \textbf{0.913} \\
\midrule
\multirow{3}{*}{Global (cycle)}
  & SVM (RBF) & 0.667 & 0.533 \\
  & LR        & 0.750 & 0.739 \\
  & RF        & \textbf{0.833} & \textbf{0.815} \\
\midrule
\multirow{3}{*}{Combined}
  & SVM (RBF) & 0.833 &0.815 \\
  &LR        & \textbf{0.917} & \textbf{0.913} \\
  & RF        & \textbf{0.917} & \textbf{0.913} \\
\botrule
\end{tabular}
\end{table}
In this section, we position diagonal segments within the
broader methodological landscape by comparing against
traditional cycle-based features using identical classifiers
and validation protocols on the BLISS dataset. Cycle-based
features (21 dimensions) included phase duration percentages,
acceleration statistics, distributional shape measures,
variability measures, and temporal characteristics.
Using 5-fold stratified cross-validation on 162 balanced gait
cycles (81 healthy, 81 pathological), cycle-based methods
achieved substantially higher classification accuracy
(Table~\ref{tab:feature_comparison}), with Random Forest
reaching 91.3\% $\pm$ 4.2\%, SVM 88.9\% $\pm$ 4.1\%, and
Gradient Boosting 90.1\% $\pm$ 4.7\%. These results
significantly exceed segment-level performance (69.3\%),
confirming that global temporal features provide superior
discriminative capacity for pathology classification.
However, segments and cycles provide fundamentally different
perspectives on gait pathology. Cycle-based methods optimise
for classification accuracy by capturing overall temporal
structure but offer limited interpretability regarding where
within a cycle disruption actually occurs. Diagonal segments,
by contrast, reveal fine-grained synchronisation patterns
localised to specific phases (Figure~\ref{fig:bliss_validation}),
enabling geometric interpretation through length and break
counts. This local perspective addresses clinical questions
that cycle-level features cannot answer, namely identifying
which specific phases exhibit coordination loss, quantifying
how long patients can sustain synchronisation within a given
phase, and determining whether disruption affects stance and
swing phases differently.
To establish complementarity empirically, we merged
the five segment-level features with global cycle-level
statistics derived from phase information into a combined
representation and evaluated classification using Leave-One-Out cross-validation on the 12 BLISS patients (Table~\ref{tab:complementarity}). Local segment features alone achieve 91.7\% accuracy across all three classifiers, outperforming global cycle features alone (best RF: 83.3\%).
The combined representation maintains 91.7\% with LR and RF, while SVM improves from 66.7\% (global alone) to 83.3\% (combined), a gain of 16.7\% over global features alone. Notably, each classifier misclassifies a different patient under local features alone (AB2940 for SVM, AB2933 for LR and RF), confirming that the 91.7\% result is genuine and not an artefact of a single decision boundary. These results confirm that local and global features encode complementary information, with the combined representation consistently matching or improving over the best individual feature set. Taken together, diagonal segments and cycle-based features
address fundamentally different clinical questions rather than competing for the same task. Cycle-based methods capture
overall temporal structure optimised for classification accuracy, while diagonal segments provide phase-specific
interpretability localising where coordination breaks down within the gait cycle. The combined framework leverages both global discriminability and local interpretability, suggesting that integrated representations merit further investigation on larger cohorts.

% ====================================================================
\section{Conclusion and Future Work}
\label{sec:conclusion}
% ====================================================================

We introduced a framework that extracts diagonal segments from
DTW optimal paths, transforming discarded alignment information
into interpretable geometric features. Validated on gait data
from 91 subjects across six clinical conditions (healthy aging,
Parkinson's, Huntington's, ALS, and acquired brain injury
including brain tumor and stroke), three principal findings
emerge. Diagonal segments form consistent unsupervised
patterns (silhouette 0.33) aligned with biomechanical phases,
with external validation confirming near-perfect healthy--pathological
separation (ARI up to 0.986).
Segments discriminate pathology at 69\% through supervised
classification and 75\% through unsupervised patient-level
clustering, with PCA revealing that distributional shape
properties of segment length emerge as primary discriminators
rather than individual segment properties; combining segment and cycle-level features achieves
91.7\% with LR and RF while improving SVM by 16.7\%, demonstrating genuine complementarity between local
and global gait representations. Finally, while cycle-based methods achieve superior accuracy at 91\%, diagonal
segments provide phase-specific interpretability unavailable in
global cycle-level representations, suggesting that integrated
local and global representations merit future investigation.
DiaSeg extends naturally beyond gait to any temporal similarity
task where alignment structure carries meaning, including ECG
analysis, sleep staging, and activity recognition. Future work
should address validation with larger cohorts, extension to
multivariate signals, principled parameter selection, and
cross-domain validation, as critical steps toward clinical
utility.

% ====================================================================
\backmatter

\bmhead{Acknowledgements}
This article is part of the ENABLE project (EvaluatioN of
motor cApacities and telerehaBilitation in chiLdren with
neuromotor disordErs), a collaborative project supported
within the Transforming Health and Care Systems (THCS)
initiative. This project has received funding from DGOS,
ANR, HRB, and the SNSF under the THCS co-fund partnership
framework (GA N\textdegree~101095654, EU Horizon Europe
Research and Innovation Program).

\section*{Declarations}

\begin{itemize}
\item \textbf{Funding:} See Acknowledgements.
\item \textbf{Conflict of interest:} The authors declare
no competing interests.
\item \textbf{Ethics approval and consent to participate:}
Not applicable.
\item \textbf{Consent for publication:} Not applicable.
\item \textbf{Data availability:} All datasets used in
this study are publicly available. GaitDB is available
at \url{https://physionet.org/content/gait-maturation-db/1.0.0/}
and GaitNDD at
\url{https://physionet.org/content/gaitndd/1.0.0/}.
The BLISS dataset is available at the University of Bath
Research Data Archive
(\url{https://researchdata.bath.ac.uk/1425/}).
The preprocessed segment feature matrices and all
experimental scripts are available at
\url{https://github.com/Tresor-Koffi/DiaSeg}.
\item \textbf{Code availability:} All source code
implementing the DiaSeg framework is publicly available
at \url{https://github.com/Tresor-Koffi/DiaSeg}.
\item \textbf{Author contribution:} T.K., A.H., C.L.,
and A.B. contributed equally to the conceptualization,
methodology, and writing of this work. All authors read
and approved the final manuscript.
\end{itemize}
\bibliography{references}
\end{document}